\documentclass[11pt,letterpaper]{article}
\usepackage{emnlp2017}
\usepackage{times}
\usepackage{latexsym}
\usepackage{multirow}
\usepackage[utf8]{inputenc}
\usepackage[T1]{fontenc}
\usepackage{bm}
\usepackage{amsmath,amsfonts,amssymb}
\usepackage{graphicx}
\usepackage{xcolor,colortbl}
\usepackage{booktabs}
\usepackage{array,multirow}

\usepackage{todonotes} 

\emnlpfinalcopy

\def\emnlppaperid{737}

\title{Learning to select data for transfer learning with Bayesian Optimization}
	
\author
{
	\begin{tabular}{ccc}
	Sebastian Ruder$^{\spadesuit\clubsuit}$ & Barbara Plank$^\heartsuit$ \\
	\end{tabular}
	\\
    $^\heartsuit$Center for Language and Cognition, University of Groningen, The Netherlands\\
    $^\spadesuit$Insight Research Centre, National University of Ireland, Galway\\
    $^\clubsuit$Aylien Ltd., Dublin, Ireland\\
	{\tt \small{sebastian@ruder.io,b.plank@rug.nl}}
}

\date{}

\begin{document}

\maketitle

\begin{abstract}
Domain similarity measures can be used to gauge adaptability and select suitable data for transfer learning, but existing approaches define ad hoc measures that are deemed suitable for respective tasks. Inspired by work on curriculum learning, we propose to \textit{learn} data selection measures using Bayesian Optimization and evaluate them across models, domains and tasks.
Our learned measures outperform existing domain similarity measures significantly on three tasks: sentiment analysis, part-of-speech tagging, and parsing.  We show the importance of complementing similarity with diversity, and that learned measures are---to some degree---transferable across models, domains, and even tasks.
\end{abstract}

\section{Introduction}

Natural Language Processing (NLP) models suffer considerably when applied in the wild. The distribution of the test data is typically very different from the data used during training, 
causing a model's performance to deteriorate substantially. Domain adaptation is a prominent approach to transfer learning that can help to bridge this gap; 
many approaches were suggested so far~\cite{Blitzer2007,DaumeIII2007a,Jiang2007,Ma2014,Schnabel2014}. However, most work focused on one-to-one scenarios. 
Only recently research considered using multiple sources. Such studies are rare and typically rely on specific model transfer approaches~\cite{Mansour2009a,Wu2016a}.

Inspired by work on curriculum learning \cite{Bengio2009,Tsvetkov2016}, we instead propose---to the best of our knowledge---the first model-agnostic \textit{data selection} approach to transfer learning. 
Contrary to curriculum learning that aims at speeding up learning (see \S \ref{sec:relwork}), we aim at \textit{learning to select} the most relevant data from multiple sources using data metrics.  
While several measures have been proposed in the past~\cite{Moore2010,Axelrod2011,VanAsch2010,Plank2011,Remus2012}, prior work is limited in studying metrics mostly in isolation, using  only the notion 
of similarity~\cite{Ben-David2007} and focusing on a single task (see \S\ref{sec:relwork}). Our hypothesis is that different tasks or even different domains demand different notions 
of similarity. In this paper we go beyond prior work by i) studying a range of similarity metrics, including diversity; and ii) testing the robustness of the learned weights 
across models (e.g., whether a more complex model can be approximated with a simpler surrogate), domains and tasks (to delimit the transferability of the learned weights). 

The contributions of this work are threefold. First, we present the first model-independent approach to \textit{learn} a data selection measure for transfer learning. 
It outperforms baselines across three tasks and multiple domains and is competitive with state-of-the-art domain adaptation approaches. 
Second, prior work on transfer learning mostly focused on similarity. We demonstrate empirically that diversity is as important as---and complements---domain similarity 
for transfer learning. Finally, we show---for the first time---to what degree learned measures \textit{transfer} across models, domains and tasks. 

\section{Background: Transfer learning}

Transfer learning generally involves the concepts of a domain and a task \cite{Pan2010}. A domain $\mathcal{D}$ consists of a feature space $\mathcal{X}$ and a marginal probability distribution $P(X)$ over $\mathcal{X}$, where $X = \{x_1, \cdots, x_n\} \in \mathcal{X}$. For document classification with a bag-of-words, $\mathcal{X}$ is the space of all document vectors, $x_i$ is the $i$-th document vector, and $X$ is a sample of documents.

Given a domain $\mathcal{D} = \{\mathcal{X},P(X)\}$, a task $\mathcal{T}$ consists of a label space $\mathcal{Y}$ and a conditional probability distribution $P(Y | X)$ that is typically learned from training data consisting of pairs $\{x_i,y_i\}$, where $x_i \in X$ and $y_i \in Y$.

Finally, given a source domain $\mathcal{D}_S$, a corresponding source task $\mathcal{T}_S$, as well as a target domain $\mathcal{D}_T$ and a target task $\mathcal{T}_T$, transfer learning seeks to facilitate the learning of the target conditional probability distribution $P(Y_T|X_T)$ in $\mathcal{D}_T$ with the information gained from $\mathcal{D}_S$ and $\mathcal{T}_S$ where $\mathcal{D}_S \neq \mathcal{D}_T$ or $\mathcal{T}_S \neq \mathcal{T}_T$. We will focus on the scenario where $\mathcal{D}_S \neq \mathcal{D}_T$ assuming that $\mathcal{T}_S = \mathcal{T}_T$, commonly referred to as domain adaptation. We investigate transfer across tasks in \textsection \ref{sec:task-transfer}.

Existing research in domain adaptation has generally focused on the scenario of one-to-one adaptation: Given a set of source domains $A$ and a set of target domains $B$, a model is evaluated based on its ability to adapt between all pairs $(a,b)$ in the Cartesian product $A \times B$ where $a \in A$ and $b \in B$ \cite{Remus2012}. However, adaptation between two dissimilar domains is often undesirable, as it may lead to negative transfer \cite{Rosenstein2005}. Only recently, many-to-one adaptation \cite{Mansour2009a,Wu2016a} has received some attention, as it replicates the realistic scenario of multiple source domains where performance on the target domain is the foremost objective.

\section{Data selection model}

In order to select training data for adaptation for a task $\mathcal{T}$, existing approaches rank the available $n$ training examples $X = \{x_1, x_2, \cdots, x_n \}$ of $k$ source domains $D = \{ \mathcal{D}_1, \mathcal{D}_2, \cdots, \mathcal{D}_k \}$ according to a domain similarity measure $\mathcal{S}$ and choose the top $m$ samples for training their algorithm. While this has been shown to work empirically \cite{Moore2010,Axelrod2011,Plank2011,VanAsch2010,Remus2012}, using a pre-existing metric leaves us unable to adapt to the characteristics of our task $\mathcal{T}$ and target domain $\mathcal{D}_T$ and foregoes additional knowledge that may be gleaned from the interaction of different metrics. For this reason, we propose to \emph{learn} the following linear domain similarity measure $\mathcal{S}$ as a linear combination of feature values:

\begin{equation} \label{eq:learned-measure}
\mathcal{S} = \phi(X) \cdot w^\intercal 
\end{equation}

where $\phi(X) \in \mathbb{R}^{n \times l}$ are the similarity and diversity features further described in \textsection \ref{sec:features} for each training example, with $l$ being the number of features, while $w \in \mathbb{R}^l$ are the weights learned by Bayesian Optimization.

We aim to learn weights $w$ in order to optimize the objective function $J$ of the respective task $\mathcal{T}$ on a small number of validation examples of the corresponding target domain $\mathcal{D}_T$.

\subsection{Bayesian Optimization for data selection}

As the learned measure $\mathcal{S}$ should be agnostic of the particular objective function $J$, we cannot use gradient-based methods for optimization. Similar to \citeauthor{Tsvetkov2016} \shortcite{Tsvetkov2016}, we use Bayesian Optimization \cite{Brochu2010}, which has emerged as an efficient framework to optimize any function. For instance, it has repeatedly found better settings of neural network hyperparameters than domain experts \cite{Snoek2012}. 

Given a black-box function $f : \mathbb{X} \rightarrow \mathbb{R}$, Bayesian Optimization aims to find an input $\hat{x} \in \arg \min_{x \in \mathbb{X}}f(x)$ that globally minimizes $f$. For this, it requires a prior $p(f)$ over the function and an acquisition function $a_{p(f)} : \mathbb{X} \rightarrow \mathbb{R}$ that calculates the \emph{utility} of any evaluation at any $x$.

Bayesian Optimization then proceeds iteratively. At iteration $t$, 1) it finds the most promising input $x_t \in \arg \max a_p(x)$ through numerical optimization; 2) it then evaluates the surrogate function $y_t \sim f(x_t) + \mathcal{N}(0, \sigma^2)$ on this input and adds the resulting data point $(x_t, y_t)$ to the set of observations $\mathcal{O}_{t-1} = (x_j, y_j)_{j=1...t-1}$; 3) finally, it updates the prior $p(f | \mathcal{O}_t)$ and the acquisition function $a_{p(f | \mathcal{O}_t)}$.

For data selection, the black-box function $f$ looks as follows: 1) It takes as input a set of weights $w$ that should be evaluated; 2) the training examples of all source domains are then scored and sorted according to Equation \ref{eq:learned-measure}; 3) the model for the respective task $\mathcal{T}$ is trained on the top $n$ samples; 4) the model is evaluated on the validation set according to the evaluation measure $J$ and the value of $J$ is returned.

Gaussian Processes (GP) are a popular choice for $p(f)$ due to their descriptive power \cite{Rasmussen2006}. We use GP with Monte Carlo acquistion and Expected Improvement (EI) \cite{Mockus1974} as acquisition function as this combination has been shown to outperform comparable approaches \cite{Snoek2012}.\footnote{We also experimented with \texttt{FABOLAS} \cite{Klein2017}, but found its ability to adjust the training set size during optimization to be inconclusive for our relatively small training sets.}

\subsection{Features} \label{sec:features}

Existing work on data selection for domain adaptation selects data based on its \textit{similarity} to the target domain. Several measures have been proposed in the literature \cite{VanAsch2010,Plank2011,Remus2012}, but were so far only used in isolation.  

Only selecting training instances with respect to the target domain also fails to account for instances that are richer and better suited for knowledge acquisition. For this reason, we consider---to our knowledge for the first time---whether intrinsic qualities of the training data accounting for \textit{diversity} are of use for domain adaptation in NLP.

\paragraph{Similarity} $\:$ We use a range of similarity metrics. Some metrics might be better suited for some tasks, while different measures might capture complementary information. We thus use the following measures as features for learning a more effective domain similarity metric.

We define similarity features over probability distributions in accordance with existing literature \cite{Plank2011}. Let $P$ be the representation of a source training example, while $Q$ is the corresponding target domain representation. Let further $M = \frac{1}{2} (P + Q)$, i.e. the average distribution between $P$ and $Q$ and let $D_{KL}(P||Q) = \sum_{i=1}^n p_i \log \frac{p_i}{q_i}$, i.e., the KL divergence between the two domains. We do not use $D_{KL}$ as a feature as it is undefined for distributions where some event $q_i \in Q$ has probability $0$, which is common for term distributions. Our features are:

\begin{itemize}
\itemsep-0.3em 
	\item Jensen-Shannon divergence \cite{Lin1991}: \\$\frac{1}{2} [D_{KL}(P||M) + D_{KL}(Q||M)]$. Jensen-Shannon divergence is a smoothed, symmetric variant of $D_{KL}$ that has been successfully used for domain adaptation \cite{Plank2011,Remus2012}.
	\item Rényi divergence \cite{Renyi1961}: \\$\frac{1}{\alpha-1} \: \log (\sum_{i=1}^n \frac{p_i^\alpha}{q_i^{\alpha-1}})$. Rényi divergence reduces to $D_{KL}$ if $\alpha=1$. We set $\alpha=0.99$ following \citeauthor{VanAsch2010} \shortcite{VanAsch2010}.
	\item Bhattacharyya distance \cite{Bhattacharya1943}: $\ln(\sum_i \sqrt{P_i Q_i})$
	\item Cosine similarity \cite{Lee2001}: $\frac{P \cdot Q}{ \|P\| \: \|Q\|} $. We can treat the distributions alternatively as vectors and consider geometrically motivated distance functions such as cosine similarity as well as the following.
	\item Euclidean distance \cite{Lee2001}: $\sqrt{\sum_i (P_i-Q_i)^2}$.
	\item Variational dist.\ \cite{Lee2001}: $\sum_i | P_i - Q_i |$.
\end{itemize}

We consider three different representations for calculating the above domain similarity measures:

\begin{itemize}
\itemsep-0.3em
	\item Term distributions \cite{Plank2011}: $t \in \mathbb{R}^{|V|}$ where $t_i$ is the probability of the $i$-th word in the vocabulary $V$. 
	\item Topic distributions \cite{Plank2011}: $t \in \mathbb{R}^{n}$ where $t_i$ is the probability of the $i$-th topic as determined by an LDA model \cite{Blei2003} trained on the data and $n$ is the number of topics.
	\item Word embeddings \cite{Mikolov2013d}: $\frac{1}{n} \sum_i v_{w_i} \sqrt{\frac{a}{p(w_i)}} $ where $n$ is the number of words with embeddings in the document, $v_{w_i}$ is the pre-trained embedding of the $i$-th word, $p(w_i)$ its probability, and $a$ is a smoothing factor used to discount frequent probabilities. A similar weighted sum has recently been shown to outperform supervised approaches for other tasks \cite{Arora2017}. As embeddings may be negative, we use them only with the latter three geometric features above.
\end{itemize}

\paragraph{Diversity} For each training example, we calculate its diversity based on the words in the example. Let $p_i$ and $p_j$ be probabilities of the word types $t_i$ and $t_j$ in the training data and $\cos(v_{t_i}, v_{t_j})$ the cosine similarity between their word embeddings. We employ measures that have been used in the past for measuring diversity \cite{Tsvetkov2016}: 

\begin{itemize}
\itemsep-0.3em 
	\item Number of word types: $\#types$.
	\item Type-token ratio: $\frac{\#types}{\#tokens}$.
	\item Entropy \cite{Shannon1948}: $- \sum_i p_i \ln (p_i)$.
	\item Simpson's index \cite{Simpson1949}: $- \sum_i p_i^2$.
	\item Rényi entropy \cite{Renyi1961}: \\$\frac{1}{\alpha-1} \: \log(\sum_i p_i^\alpha)$
	\item Quadratic entropy \cite{Rao1982}: \\$\sum_{i,j} \cos(v_{t_i}, v_{t_j}) p_i p_j$.
\end{itemize}

\section{Experiments}

\subsection{Tasks, datasets, and models}

We evaluate our approach on three tasks: sentiment analysis, part-of speech (POS) tagging, and dependency parsing. 
We use the $n$ examples with the highest score as determined by the learned data selection measure for training our models.\footnote{All code is available at \url{https://github.com/sebastianruder/learn-to-select-data}.} We show statistics for all datasets in Table \ref{tab:data-stats}.

\paragraph{Sentiment Analysis} For sentiment analysis, we evaluate on the Amazon reviews dataset \cite{Blitzer2006}.  We use tf-idf-weighted unigram and bigram features and a linear SVM classifier \cite{Blitzer2007}. We set the vocabulary size to 10,000 and the number of training examples $n=1600$ to conform with existing approaches \cite{Bollegala2011} and stratify the training set.

\paragraph{POS tagging} For POS tagging and parsing, we evaluate on the coarse-grained POS data (12 universal POS) of the SANCL 2012 shared task \cite{Petrov2012}. Each domain---except for WSJ---contains around 2000-5000 labeled sentences and more than 100,000 unlabeled sentences. In the case of WSJ, we use its dev and test data as labeled samples and treat the remaining sections as unlabeled. We set $n=2000$ for POS tagging and parsing to retain enough examples for the most-similar-domain baseline.

To evaluate the impact of model choice, we compare two models: a Structured Perceptron (in-house implementation with commonly used features pertaining to tags, words, case, prefixes, as well as prefixes and suffixes) trained for 5 iterations with a learning rate of 0.2; and a state-of-the-art Bi-LSTM tagger \cite{Plank2016a} with word and character embeddings as input. We perform early stopping on the validation set with patience of 2 and use otherwise default hyperparameters\footnote{\url{https://github.com/bplank/bilstm-aux}} as provided by the authors.

\paragraph{Parsing} For parsing, we evaluate the state-of-the-art Bi-LSTM parser by \citeauthor{Kiperwasser2016} \shortcite{Kiperwasser2016} with default hyperparameters.\footnote{\url{https://github.com/elikip/bist-parser}} We use the same domains as used for POS tagging, i.e., the dependency parsing data with gold POS as made available in the SANCL 2012 shared task.\footnote{We leave investigating the effect of the adapted taggers on parsing for future work.}

\begin{table}[]
\centering
\begin{tabular}{l l c c}
\toprule
$\mathcal{T}$ & \textbf{Domain} & \textbf{\# labeled} & \textbf{\# unlabeled} \\
\midrule
\multirow{4}{*}{\rotatebox[origin=c]{90}{Sentiment}} & Book & 2000 & 4465\\
& DVD & 2000 & 3586 \\
& Electronics & 2000 & 5681\\
& Kitchen & 2000 & 5945\\
\midrule
%
%
\multirow{6}{*}{\rotatebox[origin=c]{90}{POS/Parsing}} & Answers & 3489 & 27274\\
 & Emails & 4900 & 1194173 \\
& Newsgroups & 2391 & 1000000 \\
& Reviews & 3813 & 1965350 \\
& Weblogs & 2031 & 524834 \\
& WSJ & 2976 & 30060\\
\bottomrule
\end{tabular}
\caption{Number of labeled and unlabeled sentences for each domain in the Amazon Reviews dataset \cite{Blitzer2006} (above) and the SANCL 2012 dataset \cite{Petrov2012} for POS tagging and parsing (below).}
\label{tab:data-stats}
\end{table}

\begin{table*}[]
\centering
\resizebox{\textwidth}{!}{%
\begin{tabular}{l l l l l l}
& & \multicolumn{4}{c}{\cellcolor[gray]{.85}\textbf{Target domains}} \\
& \textbf{Feature set}  & \textbf{Book} & \textbf{DVD} & \textbf{Electronics} & \textbf{Kitchen} \\\hline
\multirow{3}{*}{\rotatebox[origin=c]{90}{Base}} & Random & 71.17 ($\pm$ 4.41) & 70.51 ($\pm$ 3.33) & 76.75 ($\pm$ 1.77) & 77.94 ($\pm$ 3.72) \\
& Jensen-Shannon divergence -- examples & 72.51 ($\pm$ 0.42) & 68.21 ($\pm$ 0.34) & 76.51 ($\pm$ 0.63) & 77.47 ($\pm$ 0.44) \\
& Jensen-Shannon divergence -- domain & 75.26 ($\pm$ 1.25) & 73.74 ($\pm$ 1.36) & 72.60 ($\pm$ 2.19)  & 80.01 ($\pm$ 1.93) \\\hline
\multirow{8}{*}{\rotatebox[origin=c]{90}{Learned measures}}  & Similarity (word embeddings) & 75.06 ($\pm$ 1.38) & 74.96 ($\pm$ 2.12) & 80.79 ($\pm$ 1.31) & 83.45 ($\pm$ 0.96)\\
& Similarity (term distributions) & 75.39 ($\pm$ 0.98) & 76.25 ($\pm$ 0.96) & 81.91 ($\pm$ 0.57) & 83.39 ($\pm$ 0.84)\\
& Similarity (topic distributions) & 76.04 ($\pm$ 1.10) & 75.89 ($\pm$ 0.81) & 81.69 ($\pm$ 0.96) & 83.09 ($\pm$ 0.95)\\
& Diversity & 76.03 ($\pm$ 1.28) & 77.48 ($\pm$ 1.33) & 81.15 ($\pm$ 0.67) & 83.94 ($\pm$ 0.99)\\
& Sim (term dists) + sim (topic dists) & 75.76 ($\pm$ 1.30) & 76.62 ($\pm$ 0.95) & 81.73 ($\pm$ 0.63) & 83.43 ($\pm$ 0.75)\\
& Sim (word embeddings) + diversity & 75.52 ($\pm$ 0.98) & 77.50 ($\pm$ 0.61) & 80.97 ($\pm$ 0.83) & 84.28 ($\pm$ 1.02)\\
& Sim (term dists) + diversity & \underline{76.20} ($\pm$ 1.45) & \underline{77.60} ($\pm$ 1.01) & \textbf{82.67} ($\pm$ 0.73) & \textbf{84.98} ($\pm$ 0.60)\\
& Sim (topic dists) + diversity & \textbf{77.16} ($\pm$ 0.77) & \textbf{79.00} ($\pm$ 0.93) & \underline{81.92} ($\pm$ 1.32) & \underline{84.29} ($\pm$ 1.00)\\
\hline
 & All source data (6,000 training examples) & 70.86 ($\pm$ 0.51) & 68.74 ($\pm$ 0.32) & 77.39 ($\pm$ 0.32) & 73.09 ($\pm$ 0.41)\\
\end{tabular}
}
\caption{Accuracy scores for data selection for sentiment analysis domain adaptation on the Amazon reviews dataset \cite{Blitzer2006}. Best: bold; second-best: underlined.}
\label{tab:sentiment-analysis-results}
\end{table*}

\subsection{Training details}

In practice, as feature values occupy different ranges, we have found it helpful to apply $z$-normalisation similar to \citeauthor{Tsvetkov2016} \shortcite{Tsvetkov2016}. We moreover constrain the weights $w$ to $[-1,1]$. 

For each dataset, we treat each domain as target domain and all other domains as source domains. Similar to \citeauthor{Bousmalis2016} \shortcite{Bousmalis2016}, we chose to use a small number (100) target domain examples as validation set. We optimize each similarity measure using Bayesian Optimization with 300 iterations according to the objective measure $J$ of each task (accuracy for sentiment analysis and POS tagging; LAS for parsing) with respect to the validation set of the corresponding target domain.

Unlabeled data is used in addition to calculate the representation of the target domain and to calculate the source domain representation for the most similar domain baseline. We train an LDA model \cite{Blei2003} with 50 topics and 10 iterations for topic distribution-based representations and use GloVe embeddings \cite{Pennington2014} trained on 42B tokens of Common Crawl data\footnote{\url{https://nlp.stanford.edu/projects/glove/}} for word embedding-based representations. 

For sentiment analysis, we conduct 10 runs of each feature set for every domain and report mean and variance. For POS tagging and parsing, we observe that variance is low and perform one run while retaining random seeds for reproducibility.

\subsection{Baselines and features}

We compare the learned measures to three baselines: i) a random baseline that randomly selects $n$ training samples from all source domains (\emph{random}); ii) the top $n$ examples selected using Jensen-Shannon divergence (\emph{JS -- examples}), which outperformed other measures in previous work \cite{Plank2011,Remus2012}; iii) $n$ examples randomly selected from the most similar source domain determined by Jensen-Shannon divergence (\emph{JS -- domain}). We additionally compare against training on all available source data (6,000 examples for sentiment analysis; 14,700-17,569 examples for POS tagging and parsing depending on the target domain).

We optimize data selection using Bayesian Optimization with every feature set: similarity features respectively based on i) word embeddings, ii) term distributions, and iii) topic distributions; and iv) diversity features. In addition, we investigate how well different representations help each other by using similarity features with the two best-performing representations, term distributions and topic distributions. Finally, we explore whether diversity and similarity-based features complement each other by in turn using each similarity-based feature set together with diversity features.

\section{Results}

\begin{table*}[]
\centering
\resizebox{\textwidth}{!}{%
\begin{tabular}{l l |c c c| c c c| c c c| c c c| c c c| c c c}
& \textbf{Trg domains $\rightarrow$} & \multicolumn{3}{c|}{\textbf{Answers}} & \multicolumn{3}{c|}{\textbf{Emails}} & \multicolumn{3}{c|}{\textbf{Newsgroups}} & \multicolumn{3}{c|}{\textbf{Reviews}} & \multicolumn{3}{c|}{\textbf{Weblogs}} & \multicolumn{3}{c}{\textbf{WSJ}}\\
& \textbf{Task $\rightarrow$} & \multicolumn{2}{c}{\textbf{POS}} & \textbf{Pars} & \multicolumn{2}{c}{\textbf{POS}} & \textbf{Pars} & \multicolumn{2}{c}{\textbf{POS}} & \textbf{Pars} & \multicolumn{2}{c}{\textbf{POS}} & \textbf{Pars} & \multicolumn{2}{c}{\textbf{POS}} & \textbf{Pars} & \multicolumn{2}{c}{\textbf{POS}} & \textbf{Pars} \\
& \textbf{Feat $\downarrow \:$ Model $\rightarrow$} & \textbf{P} & \textbf{B} & \textbf{BIST} & \textbf{P} & \textbf{B} & \textbf{BIST} & \textbf{P} & \textbf{B} & \textbf{BIST} & \textbf{P} & \textbf{B} & \textbf{BIST} & \textbf{P} & \textbf{B} & \textbf{BIST} & \textbf{P} & \textbf{B} & \textbf{BIST} \\\hline
\multirow{3}{*}{\rotatebox[origin=c]{90}{Base}}& Random & 91.34  & 92.55 & 81.02 & 91.80 & 93.25 & 79.09 & 92.50 & 93.26 & 80.61 & 92.08 & 92.12 & 82.30 & 92.76 & 93.03 & 82.39 & 91.08 & 92.54 & 78.31 \\
& JS -- examples & 92.42 & 93.16 & 82.80 & 91.75 & 93.77 & 80.53 & 92.96 & \underline{94.29} & 83.25 & 92.77 & 93.32 & 84.35 & 94.33 & 94.92 & 85.36 & 92.85 & 94.08 & 82.43\\
& JS -- domain & 90.84 & 91.13 & 80.37 & 91.64 & 93.16 & 79.93 & 92.23 & 92.67 & 81.77 & 92.27 & 92.67 & 82.11 & 93.19 & 94.34 & 83.44 & 91.20 & 92.99 & 80.61\\
\hline
\multirow{8}{*}{\rotatebox[origin=c]{90}{Learned measures}} & W2v sim & 92.53 & 93.22 & 82.74 & 92.94  & 94.14 & 81.18 & 93.41 & 94.09 & 81.62 & 93.51 & 93.30 & 82.98 & 94.41 &  94.83 & 84.30 & 93.02 & \underline{94.66} & 81.57\\
& Term sim  & \textbf{93.13} & 93.43 & \textbf{83.79} & 92.96 & 94.04 & 81.09 & 93.58 & \textbf{94.55} & 82.68 & \underline{93.53} & \textbf{93.73} & \textbf{84.66} & 94.42 & \textbf{95.09} & 84.85 & \textbf{93.44} & 94.11 & 82.57 \\
& Topic sim  & 92.50 & 93.16 & 82.87 & 92.70 & \textbf{94.48} & \underline{81.43} & 93.97 & 94.09 & 82.07 & 93.21 & 93.22 & 83.98 & 94.42 & 93.71 & 84.98 & 93.09 & 94.02 & \textbf{82.90}\\ 
& Diversity & 92.33 & 92.58 & 83.01 & \underline{93.08} & 93.56 & 80.93 & \textbf{94.37} & 93.97 & 83.98 & 93.33 & 93.05 & 83.92 & \underline{94.62} & 94.94 & \underline{85.84} & 93.33 & 93.44 & 82.80\\
& Term+topic sim  & 92.80 & \textbf{93.69} & 82.87 & 92.70 & 92.28 & 81.13 & 93.57 & 93.76 & 82.97 & \textbf{93.56} & \underline{93.61} & \underline{84.65} & 94.41 & 94.23 & 84.43 & 93.07 & \textbf{94.68} & 82.43\\ 
& W2v sim+div & 92.76 & 92.38 & 82.34 & \textbf{93.51} & \underline{94.19} & 80.77 & 93.96 & 94.10 & \textbf{84.26} & 93.45 & 93.39 & 84.47 & 94.36 & 94.95 & 85.53 & 93.32 & 93.20 & 82.32\\
& Term sim+div & 92.73 & 93.46 & \underline{83.72} & 92.90 & 93.81 & \textbf{81.60} & \underline{94.03} & 93.47 & 82.80 & 93.47 & 93.29 & 84.62 & \textbf{94.76} & \underline{95.06} & 85.44 & 93.32 & 93.68 & \underline{82.87}\\ 
& Topic sim+div & \underline{92.93} & \underline{93.62} & 82.60 & 92.62 & 93.93 & 80.83 & 93.85 & 94.06 & \underline{84.04} & 93.16 & 93.59 & 84.45 & 94.42 & 94.45 & \textbf{85.89} & \underline{93.38} & 94.23 & 82.33\\
\hline
& All source data & 94.30 & 95.16 & 86.34 & 94.34 & 95.90 & 85.57 & 95.40 & 95.90 & 87.18 & 94.90 & 95.03 & 87.51 & 95.53 & 95.79 & 88.23 & 94.19 & 95.64 & 85.20 \\

\end{tabular}
}
\caption{Results for data selection for part-of-speech tagging and parsing domain adaptation on the SANCL 2012 shared task dataset \cite{Petrov2012}. POS: Part-of-speech tagging. Pars: Parsing. POS tagging models: Structured Perceptron (P); Bi-LSTM tagger (B) \cite{Plank2016a}. Parsing model: Bi-LSTM parser (BIST) \cite{Kiperwasser2016}. Evaluation metrics: Accuracy (POS tagging); Labeled Attachment Score (parsing). Best: bold; second-best: underlined.}
\label{tab:pos-tagging-parsing-results}
\end{table*}

\paragraph{Sentiment analysis} We show results for sentiment analysis in Table \ref{tab:sentiment-analysis-results}. First of all, the baselines show that 
the sentiment review domains are clearly delimited. Adapting between two similar domains such as Book and DVD is more productive than adaptation between dissimilar domains, e.g.\ Books and Electronics, as shown in previous work \cite{Blitzer2007}. This explains the strong performance of the most-similar-domain baseline. 
In contrast, selecting individual examples based on a domain similarity measure performs only as good as chance. Thus, when domains are more clear-cut, selecting from the closest domain is a stronger baseline than selecting from the entire pool of source data. 

If we learn a data selection measure using Bayesian Optimization, we are able to outperform the baselines with almost all feature sets. Performance gains are considerable for all domains with individual feature sets (term similarity, word embeddings similarity, diversity and topic similarity), except for Books were improvements for some single feature sets are smaller. Term distributions and topic distributions are the best-performing representations for calculating similarity, with term distributions performing slightly better across all domains. Combining term distribution-based and topic distribution-based features only provides marginal gains over the individual feature sets, demonstrating that most of the information is contained in the similarity features rather than the representations.

Diversity features perform comparatively to the best similarity features and outperform them on two domains. Furthermore, the combination of diversity and similarity features yields another sizable gain of around 1 percentage point for almost all domains over the best similarity features, which shows that diversity and similarity features capture complementary information. Term distribution and topic distribution-based similarity features in conjunction with diversity features finally yield the best performance, outperforming the baselines by 2-6 points in absolute accuracy.

Finally, we compare data selection to training on all available source data (in this setup, 6,000 instances). The result complements the findings of the most-similar baseline: as domains are dissimilar, training on all available sources is detrimental. 

 \begin{figure}[ht!]
 \includegraphics[width=1.05\columnwidth]{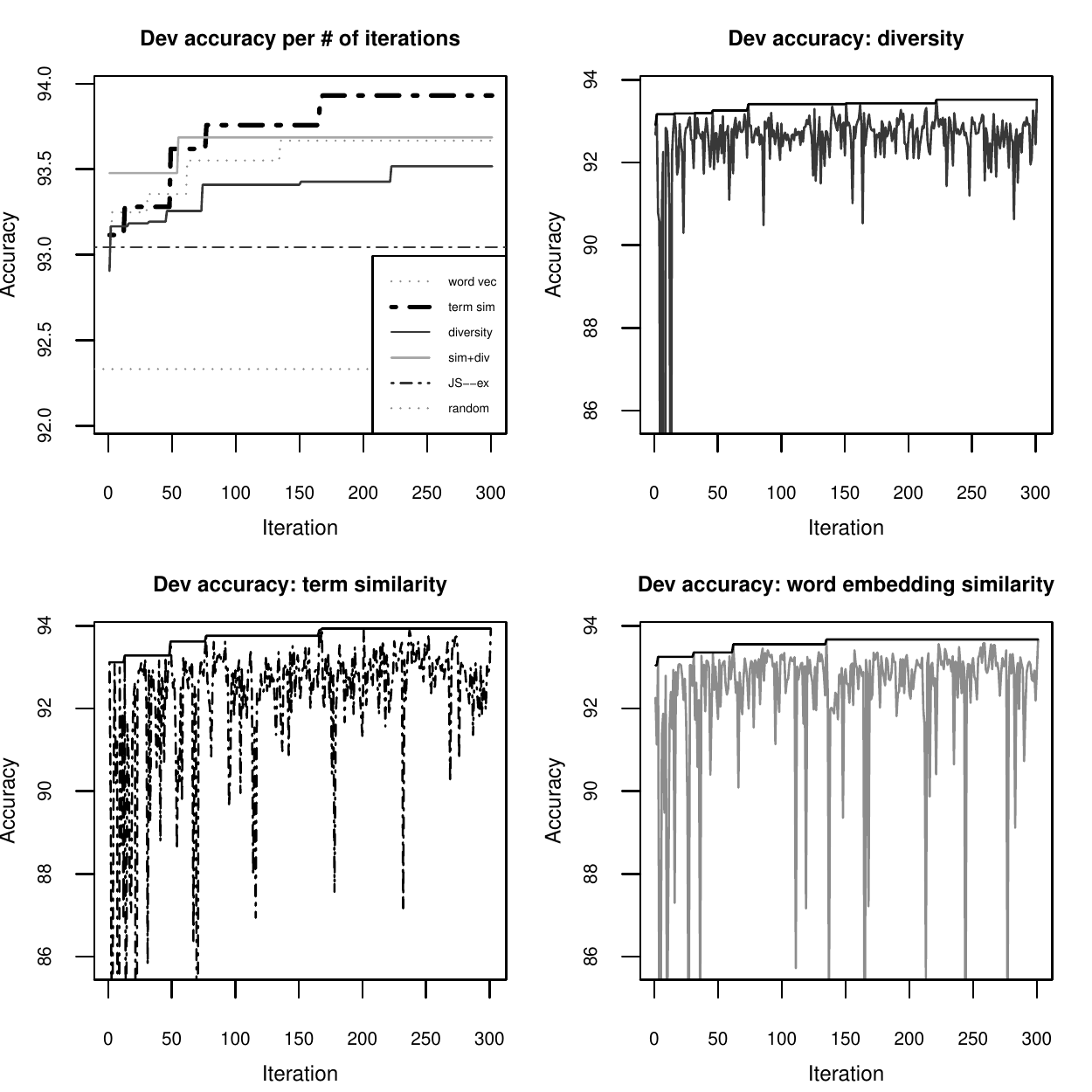}
 \caption{Dev accuracy curves of Bayes Optimization for POS tagging on the Reviews domain. Best dev acc for different feature sets (top-left). Best dev acc vs. exploration (top-right, bottom).}
  \label{fig:pos-dev-curves}
  \end{figure}

\paragraph{POS tagging} Results for POS tagging are given in Table \ref{tab:pos-tagging-parsing-results}. 
Using Bayesian Optimization, we are able to outperform the baselines with almost all feature sets, except for a few cases (e.g., diversity and word embeddings similarity, topic and term distributions). Overall term distribution-based similarity emerges as the most powerful individual feature. Combining it with diversity does not prove as beneficial as in the sentiment analysis case, however, often yields the second-best results. 

Notice that for POS tagging/parsing, in contrast to sentiment analysis, the most-similar domain baseline is not effective, it often performs only as good as chance, or even hurts. In contrast, the baseline that selects instances (\emph{JS -- examples}) rather than a domain performs better.
This makes sense as in SA topically closer domains express sentiment in more similar ways, while for POS tagging having more varied training instances is intuitively more beneficial. In fact, when inspecting the domain distribution of our approach, we find that the best SA model chooses more instances from the closest domain, while for POS tagging instances are more balanced across domains. This suggests that the Web treebank domains are less clear-cut. In fact, training a model on all sources, which is considerably more and varied data (in this setup, 14-17.5k training instances) is beneficial. This is in line with findings in machine translation~\cite{mirkin2014data}, which show that similarity-based selection works best if domains are very different. Results are thus less pronounced for POS tagging, and we leave experimenting with larger $n$ for future work.

\begin{table*}[ht!]
\centering
\resizebox{\textwidth}{!}{%
\begin{tabular}{l c c c c c c c c c c c c}
& \multicolumn{12}{c}{\cellcolor[gray]{.85} \textbf{Target domains}} \\
& \multicolumn{2}{c}{\textbf{Answers}} & \multicolumn{2}{c}{\textbf{Emails}} & \multicolumn{2}{c}{\textbf{Newsgroups}} & \multicolumn{2}{c}{\textbf{Reviews}} & \multicolumn{2}{c}{\textbf{Weblogs}} & \multicolumn{2}{c}{\textbf{WSJ}}\\
\textbf{Feature set $\downarrow$ \: $\mathcal{M}_\mathcal{S} \rightarrow$} & \textbf{B} & \textbf{P}$_{proxy}$ & \textbf{B} & \textbf{P}$_{proxy}$& \textbf{B} & \textbf{P}$_{proxy}$& \textbf{B} & \textbf{P}$_{proxy}$& \textbf{B} & \textbf{P}$_{proxy}$& \textbf{B} & \textbf{P}$_{proxy}$\\\hline
Term similarity & \underline{93.43} & \underline{93.67} & \underline{94.04} & \underline{93.88} & \underline{94.55} & 93.77 & \textbf{\underline{93.73}} & \underline{93.54} & \textbf{\underline{95.09}} & \underline{95.06} & \underline{94.11} & \underline{94.30}\\ 
Diversity & 92.58 & \underline{93.19} & 93.56 & \textbf{\underline{94.40}} & 93.97 & \textbf{\underline{94.96}} & 93.05 & \underline{93.52} & \underline{94.94} & 94.91 & 93.44 & \textbf{\underline{94.14}}\\
Term similarity+diversity & \textbf{\underline{93.46}} & \underline{93.18} & \underline{93.81} & \underline{94.29} & 93.47 & 94.28 & 93.29 & \underline{93.35} & \underline{95.06} & 94.67 & 93.68 & 93.92
\end{tabular}
}
\caption{Accuracy scores for cross-model transfer of learned data selection weights for part-of-speech tagging from Structured Perceptron (P$_{proxy}$) to Bi-LSTM tagger (B) \cite{Plank2016a} on the SANCL 2012 shared task dataset \cite{Petrov2012}. Data selection weights are learned using model $\mathcal{M}_\mathcal{S}$; Bi-LSTM tagger (B) is then trained using the learned weights. Better than baselines: underlined.}
\label{tab:pos-tagging-model-transfer-results}
\end{table*}

To gain some insight into the optimization procedure, Figure~\ref{fig:pos-dev-curves} shows the development accuracy for the Structured Perceptron for an example domain. The top-right and bottom graphs show the hypothesis space exploration of Bayesian Optimization for different single feature sets, while the top-left graph displays the overall best dev accuracy for different features. We observe again that term similarity is among the best feature sets and results in a larger explored space (more variance), in contrast to the diversity features whose development accuracy increases less and results in an overall less explored space. Exploration plots for other features/models looks similar. 


\paragraph{Parsing} The results for parsing are given in Table \ref{tab:pos-tagging-parsing-results}. Diversity features are stronger than for POS tagging and outperform the baselines for all except the Reviews domain. Similarly to POS tagging, term distribution-based similarity as well as its combination with diversity features yield the best results across most domains.

\subsection{Transfer across models}
In addition, we are interested how well the metric learned for one target domain transfers to other settings. We first investigate its ability to transfer to another model. In practice, a metric can be learned using a model that is cheap to evaluate and serves as proxy for a state-of-the-art model, in a way similar to uptraining~\cite{Petrov2010}. For this, we employ the data selection features learned using the Structured Perceptron model for POS tagging and use them to select data for the Bi-LSTM tagger. The results in Table \ref{tab:pos-tagging-model-transfer-results} indicate that cross-model transfer is indeed possible, with most transferred feature sets achieving similar results or even outperforming features learned with the Bi-LSTM. In particular, transferred diversity significantly outperforms its in-model equivalent. This is encouraging, as it allows  to learn a data selection metric using less complex models.

\begin{table}[ht!]
\centering
\resizebox{\columnwidth}{!}{%
\begin{tabular}{l c c c c c}
& & \multicolumn{4}{c}{\cellcolor[gray]{.85} \textbf{Target domains}}\\
\textbf{Feature} & \textbf{$\mathcal{D}_\mathcal{S}$} & \textbf{B} & \textbf{D} & \textbf{E} & \textbf{K} \\\hline
Sim & B & \cellcolor[gray]{.85} \textbf{75.39} & 75.22 & 80.74 & 80.41\\
Sim & D &75.30 & \cellcolor[gray]{.85}76.25 & \textbf{82.68} & 82.29 \\
Sim & E &74.55 & 76.65 & \cellcolor[gray]{.85}81.91 & 82.23 \\
Sim & K &73.64 & \textbf{76.66} & 81.09 & \cellcolor[gray]{.85}\textbf{83.39} \\\hline
Div & B &\cellcolor[gray]{.85}\textbf{76.03} & 75.16 & 80.16 & 80.01 \\
Div & D &75.68 & \cellcolor[gray]{.85}\textbf{77.48} & 65.74 & 72.48 \\
Div & E &74.69 & 76.60 & \cellcolor[gray]{.85}\textbf{81.15} & 81.97  \\
Div & K &75.03 & 76.23 & 80.71 & \cellcolor[gray]{.85}\textbf{83.94} \\\hline
Sim+div & B & \cellcolor[gray]{.85}\textbf{76.20} & 64.81 & 65.06 & 79.87\\
Sim+div & D & 74.17 & \cellcolor[gray]{.85}77.60 & \textbf{83.26}  & \textbf{85.19}\\
Sim+div & E & 74.14 & \textbf{79.32} & \cellcolor[gray]{.85}82.67 & 84.53 \\
Sim+div & K & 75.54 & 76.11 & 78.72 & \cellcolor[gray]{.85}84.98\\\hline
SDAMS & - & 78.29 & 79.13 & 84.06 & 86.29
\end{tabular}
}
\caption{Accuracy scores for cross-domain transfer of learned data selection weights on Amazon reviews \cite{Blitzer2006}. $\mathcal{D}_\mathcal{S}$: target domain used for learning metric $\mathcal{S}$. B: Book. D: DVD. E: Electronics. K: Kitchen. Sim: term distribution-based similarity. Div: diversity. Best per feature set: bold. In-domain results: gray. SDAMS \cite{Wu2016a} listed as comparison.}
\label{tab:domain-transfer-results}
\end{table}

\begin{table*}[ht!]

\centering
\resizebox{\textwidth}{!}{%
\begin{tabular}{l c c c c c c c}
& & \multicolumn{6}{c}{\cellcolor[gray]{.85} \textbf{Target domains}}\\
\textbf{Feature set} & \textbf{$\mathcal{D}_\mathcal{S}$}  & \textbf{Answers (A)} & \textbf{Emails (E)} & \textbf{Newsgroups (N)} & \textbf{Reviews (R)} & \textbf{Weblogs (W)} & \textbf{WSJ}\\\hline
Term similarity & A & \cellcolor[gray]{.85}\textbf{93.13} & 91.60 & 93.94 & \textbf{93.63} & 94.26 & 92.42\\
Term similarity & E & 92.35 & \cellcolor[gray]{.85}\textbf{92.96} & 93.42 & \textbf{93.63} & 93.75 & 92.24\\
Term similarity & N & 92.48 & 92.28 & \cellcolor[gray]{.85}93.58 & 93.35 & 93.95 & 93.00\\
Term similarity & R & 92.06 & 92.18 & 93.38 & \cellcolor[gray]{.85}93.53 & 94.26 & 91.88\\
Term similarity & W & 92.69 & 92.12 & \textbf{93.65} & 93.12 & \cellcolor[gray]{.85}\textbf{94.42} & 92.63\\
Term similarity & WSJ & 92.50 & 92.51 & 93.53 & 93.00 & 94.29 & \cellcolor[gray]{.85}\textbf{93.44} \\\hline
Diversity & A & \cellcolor[gray]{.85} 92.33 & 92.14 & 93.46 & 92.00 & 94.01 & 92.56\\ 
Diversity & E & 92.11 & \cellcolor[gray]{.85} \textbf{93.08} & 93.81 & 92.67 & 94.16 & 93.13\\
Diversity & N & \textbf{92.67} & 92.22 & \cellcolor[gray]{.85} \textbf{94.37} & 92.44 & 94.05 & 92.96\\
Diversity & R & 92.65 & 92.72 & 93.67 & \cellcolor[gray]{.85} \textbf{93.33} & 94.18 & 93.28\\
Diversity & W & 92.19 & 92.31 & 93.31 & 92.20 & \cellcolor[gray]{.85} \textbf{94.62} & 92.04\\
Diversity & WSJ & 92.26 & 92.31 & 93.75 & 92.70 & 94.32 &  \cellcolor[gray]{.85} \textbf{93.33} \\\hline
Term similarity+diversity & A & \cellcolor[gray]{.85} 92.73 & 92.63 & 93.16 & 92.58 & 93.88 & 92.23\\
Term similarity+diversity & E & 92.55 & \cellcolor[gray]{.85} 92.90 & 93.78 & 92.73 & 93.54 & 92.57\\
Term similarity+diversity & N & 92.47 & 92.27 & \cellcolor[gray]{.85} \textbf{94.03} & 92.63 & 94.30 & 93.14\\
Term similarity+diversity & R & \textbf{92.80} & \textbf{93.11} & 93.92 & \cellcolor[gray]{.85} 93.47 & 93.79 & 92.99\\
Term similarity+diversity & W & 92.61 & 92.45 & 93.44 & \textbf{93.52} & \cellcolor[gray]{.85} \textbf{94.76} & 93.26\\
Term similarity+diversity & WSJ & 91.82 & 92.37 & 93.52 & 92.63 & 94.17 & \cellcolor[gray]{.85} \textbf{93.32} 
\end{tabular}
}
\caption{Accuracy scores for cross-domain transfer of learned data selection weights for part-of-speech tagging with the Structured Perceptron model on the SANCL 2012 shared task dataset \cite{Petrov2012}. $\mathcal{D}_\mathcal{S}$: target domain used for learning metric $\mathcal{S}$. Best: bold. In-domain results: gray.}
\label{tab:pos-tagging-domain-transfer-results}
\end{table*}

\subsection{Transfer across domains}

We explore whether data selection parameters learned for one target domain transfer to other target domains. For each domain, we use the weights with the highest performance on the validation set and use them for data selection with the remaining domains as target domains. We conduct 10 runs for the best-performing feature sets for sentiment analysis and report the average accuracy scores in Table \ref{tab:domain-transfer-results} (for POS tagging, see Table \ref{tab:pos-tagging-domain-transfer-results}).

The transfer of the weights learned with Bayesian Optimization is quite robust in most cases. Feature sets like Similarity or Diversity trained on Books outperform the strong JS -- $\mathcal{D}$  baseline in all 6 cases, for Electronics and Kitchen in 4/6 cases (off-diagonals for box 2 and 3 in Table \ref{tab:domain-transfer-results}). In some cases, the transferred weights even outperform the data selection metric learned for the respective domain, such as on D->E with sim and sim+div features and by almost 2 pp on E->D.

Transferred similarity+diversity features mostly achieve higher performance than other feature sets, but the higher number of parameters runs the risk of overfitting to the domain as can be observed with two instances of negative transfer with sim+div features.

As a reference, we also list the performance of the state-of-the-art multi-domain adaptation approach \cite{Wu2016a}, which shows that task-independent data selection is in fact competitive with a task-specific, heuristic state-of-the-art domain adaptation approach. In fact our \textit{transferred} similarity+diversity feature (E->D) outperforms the state-of-the-art \cite{Wu2016a} on DVD. This is encouraging as previous work \citep{Remus2012} has shown that data selection and domain adaptation can be complementary. 

\begin{table}[ht!]
\centering
\resizebox{\columnwidth}{!}{%
\begin{tabular}{l c c c c}
& & \multicolumn{3}{c}{\cellcolor[gray]{.85} \textbf{Target tasks}}\\
\textbf{Feature set} & \textbf{$\mathcal{T}_\mathcal{S}$} & \textbf{POS} & \textbf{Pars} & \textbf{SA} \\\hline
Sim & POS & \cellcolor[gray]{.85}\underline{93.51} & 83.11  & 74.19 \\
Sim & Pars & 92.78 & \cellcolor[gray]{.85}\underline{83.27} & 72.79 \\
Sim & SA & 86.13 & 67.33 & \cellcolor[gray]{.85}\underline{79.23} \\\hline
Div & POS & \cellcolor[gray]{.85}\underline{93.51} & 83.11 & 69.78 \\
Div & Pars & \underline{93.02} & \cellcolor[gray]{.85}\underline{83.41}  & 68.45 \\
Div & SA & 90.52 & 74.68 & \cellcolor[gray]{.85}\underline{79.65} \\\hline
Sim+div & POS & \cellcolor[gray]{.85}\underline{93.54} & \underline{83.24} & 69.79 \\
Sim+div & Pars & \underline{93.11} & \cellcolor[gray]{.85}\underline{83.51} & 72.27  \\
Sim+div & SA & 89.80 & 75.17 & \cellcolor[gray]{.85}\underline{80.36}
\end{tabular}
}
\caption{Results of cross-task transfer of learned data selection weights. $\mathcal{T}_\mathcal{S}$: task used for learning metric $\mathcal{S}$. POS: Part-of-speech tagging. Pars: Parsing. SA: sentiment analysis. Accuracy scores for SA and POS; LAS Attachment Score for parsing. Models: Structured Perceptron (POS tagging); Bi-LSTM parser \cite{Kiperwasser2016} (Pars). Same features as in Table~\ref{tab:domain-transfer-results}. In-task results: gray. Better than base: underlined.}
\label{tab:task-transfer}
\end{table}

\subsection{Transfer across tasks} \label{sec:task-transfer}

We finally investigate whether data selection is task-specific or whether a metric learned on one task can be transferred to another one. 
For each feature set, we use the learned weights for each domain in the source task (for sentiment analysis, we use the best weights on the validation set; 
for POS tagging, we use the Structured Perceptron model) and run experiments with them for all domains in the target task.\footnote{E.g., for SA->POS, for each feature set, we obtain one set of weights for each of 4 SA domains, which we use to select data for the 6 POS domains, yielding $4 \cdot 6 = 24$ results.} We report the averaged accuracy scores for transfer across all tasks in Table \ref{tab:task-transfer}.

Transfer is productive between related tasks, i.e. POS tagging and parsing results are similar to those obtained with data selection learned for the particular task. We observe large drops in performance for transfer between unrelated tasks, such as sentiment analysis and POS tagging, which is expected since these are very different tasks. Between related tasks, the combination of similarity and diversity features achieves the most robust transfer and outperforms the baselines in both cases. This suggests that even in the absence of target task data, we only require data of a related task to learn a successful data selection measure.

\section{Related work}\label{sec:relwork}

Most prior work on data selection for transfer learning focuses on phrase-based machine translation. Typically language models are leveraged via perplexity or cross-entropy scoring to select target data~\cite{Moore2010,Axelrod2011,Duh2013,mirkin2014data}. A recent study investigates data selection for neural machine translation~\cite{vanderwees2017}. Perplexity was also used to select training data for dependency parsing~\cite{Sogaard2011}, but has been found to be less suitable for tasks such as sentiment analysis \cite{Ruder2017b}. In general, there are fewer studies on data selection for other tasks, e.g., constituent parsing~\cite{McClosky2010}, dependency parsing~\cite{Plank2011,Sogaard2011} and sentiment analysis~\cite{Remus2012}.
Work on predicting task accuracy is related, but can be seen as complementary~\cite{Ravi2008,VanAsch2010}.

Many domain similarity metrics have been proposed. \citeauthor{Blitzer2007} \shortcite{Blitzer2007} show that proxy $\mathcal{A}$ distance can be used to measure the adaptability between 
two domains in order to determine examples for annotation. \citeauthor{VanAsch2010} \shortcite{VanAsch2010} find that Rényi divergence outperforms other metrics in predicting 
POS tagging accuracy, while \citeauthor{Plank2011} \shortcite{Plank2011} observe that topic distribution-based representations with Jensen-Shannon divergence perform best for data selection for 
parsing. \citeauthor{Remus2012} \shortcite{Remus2012} apply Jensen-Shannon divergence to select training examples for sentiment analysis. 
Finally, \citeauthor{Wu2016a} \shortcite{Wu2016a} propose a similarity metric based on a sentiment graph. We test previously explored similarity metrics and complement them with diversity.

Very recently interest emerged in \textit{curriculum learning}~\cite{Bengio2009}. It is inspired by human active learning by providing easier examples at initial learning stages (e.g., by curriculum strategies such as growing vocabulary size). Curriculum learning employs a range of data metrics, but aims at altering the order in which the entire training data is selected, rather than \textit{selecting} data. In contrast to us, curriculum learning is mostly aimed at speeding up the learning, while we focus on learning metrics for transfer learning. Other related work in this direction include using Reinforcement Learning to learn what data to select during neural network training~\cite{Fan2017}.

There is a long history of research in adaptive data selection, with early approaches grounded in information theory using a Bayesian learning framework \cite{MacKay1992}. It has also been studied extensively as active learning \cite{El-Gamal1991}. Curriculum learning is related to active learning~\cite{Settles2012}, whose view is different: active learning aims at finding the most difficult instances to label, examples typically close to the decision boundary. Confidence-based measures are prominent, but as such are less widely applicable than our model-agnostic approach.

The approach most similar to ours is by \citeauthor{Tsvetkov2016} \shortcite{Tsvetkov2016} who use Bayesian Optimization to learn a curriculum for training word embeddings. Rather than ordering data (in their case, paragraphs), we use Bayesian Optimization for learning to \emph{select} relevant training instances that are useful for transfer learning in order to prevent negative transfer \cite{Rosenstein2005}. To the best of our knowledge there is no prior work that uses this strategy for transfer learning.

\section{Conclusion}

We propose to use Bayesian Optimization to learn data selection measures for transfer learning. Our results  outperform existing domain similarity metrics on three tasks (sentiment analysis, POS tagging and parsing), and are competitive with a state-of-the-art domain adaptation approach. More importantly, we present the first study on the transferability of such measures, showing promising results to port them across models, domains and related tasks.

\section*{Acknowledgments}

We thank the anonymous reviewers for their valuable feedback. Sebastian is supported by Irish Research Council Grant Number EBPPG/2014/30 and Science Foundation Ireland Grant Number SFI/12/RC/2289.
Barbara is supported by NVIDIA corporation and the Computing Center of the University of Groningen.

\bibliography{learn_similarity}

\begin{thebibliography}{51}
\expandafter\ifx\csname natexlab\endcsname\relax\def\natexlab#1{#1}\fi

\bibitem[{Arora et~al.(2017)Arora, Liang, and {Tengyu Ma}}]{Arora2017}
Sanjeev Arora, Yingyu Liang, and {Tengyu Ma}. 2017.
\newblock {A Simple But Tough-to-Beat Baseline for Sentence Embeddings}.
\newblock In \emph{ICLR 2017}.

\bibitem[{Axelrod et~al.(2011)Axelrod, He, and Gao}]{Axelrod2011}
Amittai Axelrod, Xiaodong He, and Jianfeng Gao. 2011.
\newblock \href
  {https://www.microsoft.com/en-us/research/publication/domain-adaptation-via-pseudo-in-domain-data-selection/}
  {Domain adaptation via pseudo in-domain data selection}.
\newblock In \emph{Proceedings of the 2011 Conference on Empirical Methods in
  Natural Language Processing}. ACM.

\bibitem[{Ben-David et~al.(2007)Ben-David, Blitzer, Crammer, and
  Pereira}]{Ben-David2007}
Shai Ben-David, John Blitzer, Koby Crammer, and Fernando Pereira. 2007.
\newblock {Analysis of representations for domain adaptation}.
\newblock \emph{Advances in Neural Information Processing Systems},
  19:137--144.

\bibitem[{Bengio et~al.(2009)Bengio, Louradour, Collobert, and
  Weston}]{Bengio2009}
Y.~Bengio, J.~Louradour, R.~Collobert, and J.~Weston. 2009.
\newblock Curriculum learning.
\newblock In \emph{International Conference on Machine Learning, {ICML}}.

\bibitem[{Bhattacharya(1943)}]{Bhattacharya1943}
Anil Bhattacharya. 1943.
\newblock {On a measure of divergence between two statistical population
  defined by their population distributions.}
\newblock \emph{Bulletin Calcutta Mathematical Society 35.99-109 (1943): 28.},
  35(99-109):28.

\bibitem[{Blei et~al.(2003)Blei, Ng, and Jordan}]{Blei2003}
David~M. Blei, Andrew~Y. Ng, and Michael~I. Jordan. 2003.
\newblock \href {https://doi.org/10.1162/jmlr.2003.3.4-5.993} {{Latent
  Dirichlet Allocation}}.
\newblock \emph{Journal of Machine Learning Research}, 3(4-5):993--1022.

\bibitem[{Blitzer et~al.(2007)Blitzer, Dredze, and Pereira}]{Blitzer2007}
John Blitzer, Mark Dredze, and Fernando Pereira. 2007.
\newblock \href {https://doi.org/10.1109/IRPS.2011.5784441} {{Biographies,
  bollywood, boom-boxes and blenders: Domain adaptation for sentiment
  classification}}.
\newblock \emph{Annual Meeting-Association for Computational Linguistics},
  45(1):440.

\bibitem[{Blitzer et~al.(2006)Blitzer, McDonald, and Pereira}]{Blitzer2006}
John Blitzer, Ryan McDonald, and Fernando Pereira. 2006.
\newblock \href {https://doi.org/10.3115/1610075.1610094} {{Domain Adaptation
  with Structural Correspondence Learning}}.
\newblock \emph{Proceedings of the 2006 Conference on Empirical Methods in
  Natural Language Processing (EMNLP '06)}, pages 120--128.

\bibitem[{Bollegala et~al.(2011)Bollegala, Weir, and Carroll}]{Bollegala2011}
Danushka Bollegala, David Weir, and John Carroll. 2011.
\newblock {Using Multiple Sources to Construct a Sentiment Sensitive Thesaurus
  for Cross-Domain Sentiment Classification}.
\newblock In \emph{Proceedings of the 49th Annual Meeting of the Association
  for Computational Linguistics: Human Language Technologies-Volume 1}, pages
  132--141.

\bibitem[{Bousmalis et~al.(2016)Bousmalis, Trigeorgis, Silberman, Krishnan, and
  Erhan}]{Bousmalis2016}
Konstantinos Bousmalis, George Trigeorgis, Nathan Silberman, Dilip Krishnan,
  and Dumitru Erhan. 2016.
\newblock \href {http://arxiv.org/abs/1608.06019} {{Domain Separation
  Networks}}.
\newblock \emph{NIPS}.

\bibitem[{Brochu et~al.(2010)Brochu, Cora, and {De Freitas}}]{Brochu2010}
E~Brochu, V~M Cora, and N~{De Freitas}. 2010.
\newblock \href {https://doi.org/1012.2599} {{A tutorial on Bayesian
  optimization of expensive cost functions, with application to active user
  modeling and hierarchical reinforcement learning}}.
\newblock In \emph{CoRR}.

\bibitem[{{Daum{\'{e}} III}(2007)}]{DaumeIII2007a}
Hal {Daum{\'{e}} III}. 2007.
\newblock \href {https://doi.org/10.1.1.110.2062} {{Frustratingly Easy Domain
  Adaptation}}.
\newblock \emph{Association for Computational Linguistic (ACL)},
  (June):256--263.

\bibitem[{Duh et~al.(2013)Duh, Neubig, Sudoh, and Tsukada}]{Duh2013}
Kevin Duh, Graham Neubig, Katsuhito Sudoh, and Hajime Tsukada. 2013.
\newblock \href {http://www.anthology.aclweb.org/P/P13/P13-2.pdf{\#}page=726}
  {{Adaptation Data Selection using Neural Language Models: Experiments in
  Machine Translation}}.
\newblock \emph{ACL-2013}, (1):678--683.

\bibitem[{El-Gamal(1991)}]{El-Gamal1991}
M.~A El-Gamal. 1991.
\newblock {The role of priors in active Bayesian learning in the sequential
  statistical decision framework}.
\newblock In \emph{Maximum Entropy and Bayesian Methods}, pages 33--38.
  Springer Netherlands.

\bibitem[{Fan et~al.(2017)Fan, Tian, Qin, Bian, and Liu}]{Fan2017}
Yang Fan, Fei Tian, Tao Qin, Jiang Bian, and Tie-Yan Liu. 2017.
\newblock \href {http://arxiv.org/abs/1702.08635} {{Learning What Data to
  Learn}}.
\newblock In \emph{Workshop track - ICLR 2017}.

\bibitem[{Jiang and Zhai(2007)}]{Jiang2007}
Jing Jiang and ChengXiang Zhai. 2007.
\newblock \href {https://doi.org/10.1145/1273496.1273558} {{Instance Weighting
  for Domain Adaptation in NLP}}.
\newblock \emph{Proceedings of the 45th Annual Meeting of the Association of
  Computational Linguistics}, (October):264--271.

\bibitem[{Kiperwasser and Goldberg(2016)}]{Kiperwasser2016}
Eliyahu Kiperwasser and Yoav Goldberg. 2016.
\newblock \href {http://arxiv.org/abs/1603.04351} {{Simple and Accurate
  Dependency Parsing Using Bidirectional LSTM Feature Representations}}.
\newblock \emph{Transactions of the Association for Computational Linguistics},
  4:313--327.

\bibitem[{Klein et~al.(2017)Klein, Falkner, and Hutter}]{Klein2017}
Aaron Klein, Stefan Falkner, and Frank Hutter. 2017.
\newblock \href {http://arxiv.org/abs/arXiv:1605.07079v1} {{Fast Bayesian
  Optimization of Machine Learning Hyperparameters on Large Datasets}}.
\newblock In \emph{Proceedings of the 20th International Conference on
  Artificial Intelligence and Statistics (AISTATS 2017)}.

\bibitem[{Lee(2001)}]{Lee2001}
Lillian Lee. 2001.
\newblock {On the Effectiveness of the Skew Divergence for Statistical Language
  Analysis}.
\newblock \emph{AISTATS (Artificial Intelligence and Statistics)}, pages
  65--72.

\bibitem[{Lin(1991)}]{Lin1991}
J~Lin. 1991.
\newblock {Divergence Measures Based on the Shannon Entropy}.
\newblock \emph{IEEE Transactions on Information theory}, 37(1):145--151.

\bibitem[{Ma et~al.(2014)Ma, Zhang, and Zhu}]{Ma2014}
Ji~Ma, Yue Zhang, and Jingbo Zhu. 2014.
\newblock \href {http://www.aclweb.org/anthology/P14-1014} {{Tagging The Web:
  Building A Robust Web Tagger with Neural Network}}.
\newblock \emph{Proceedings of the 52nd Annual Meeting of the Association for
  Computational Linguistics (Volume 1: Long Papers)}, pages 144--154.

\bibitem[{MacKay(1992)}]{MacKay1992}
David J.~C. MacKay. 1992.
\newblock \href {https://doi.org/10.1162/neco.1992.4.4.590} {{Information-Based
  Objective Functions for Active Data Selection}}.
\newblock \emph{Neural Computation}, 4(4):590--604.

\bibitem[{Mansour(2009)}]{Mansour2009a}
Yishay Mansour. 2009.
\newblock \href {http://arxiv.org/abs/arXiv:0902.3430v2} {{Domain Adaptation
  with Multiple Sources}}.
\newblock \emph{Neural Information Processing Systems Conference (NIPS 2009)}.

\bibitem[{McClosky et~al.(2010)McClosky, Charniak, and Johnson}]{McClosky2010}
David McClosky, Eugene Charniak, and Mark Johnson. 2010.
\newblock Automatic domain adaptation for parsing.
\newblock In \emph{Human Language Technologies: The 2010 Annual Conference of
  the North American Chapter of the Association for Computational Linguistics},
  pages 28--36. Association for Computational Linguistics.

\bibitem[{Mikolov et~al.(2013)Mikolov, Chen, Corrado, and Dean}]{Mikolov2013d}
Tomas Mikolov, Kai Chen, Greg Corrado, and Jeffrey Dean. 2013.
\newblock \href {http://arxiv.org/abs/1310.4546} {{Distributed Representations
  of Words and Phrases and their Compositionality}}.
\newblock \emph{NIPS}.

\bibitem[{Mirkin and Besacier(2014)}]{mirkin2014data}
Shachar Mirkin and Laurent Besacier. 2014.
\newblock Data selection for compact adapted smt models.
\newblock In \emph{Eleventh Conference of the Association for Machine
  Translation in the Americas (AMTA)}.

\bibitem[{Mo{\v{c}}kus(1974)}]{Mockus1974}
Jonas Mo{\v{c}}kus. 1974.
\newblock {On bayesian methods for seeking the extremum}.
\newblock In \emph{Optimization Techniques IFIP Technical Conference
  Novosibirsk}.

\bibitem[{Moore and Lewis(2010)}]{Moore2010}
Robert~C Moore and William~D Lewis. 2010.
\newblock {Intelligent Selection of Language Model Training Data}.
\newblock \emph{ACL-2010: 48th Annual Meeting of the Association for
  Computational Linguistics}, (July):220--224.

\bibitem[{Pan and Yang(2010)}]{Pan2010}
Sinno~Jialin Pan and Qiang Yang. 2010.
\newblock {A survey on transfer learning}.
\newblock \emph{IEEE Transactions on Knowledge and Data Engineering},
  22(10):1345--1359.

\bibitem[{Pennington et~al.(2014)Pennington, Socher, and
  Manning}]{Pennington2014}
Jeffrey Pennington, Richard Socher, and Christopher~D. Manning. 2014.
\newblock \href {https://doi.org/10.3115/v1/D14-1162} {{Glove: Global Vectors
  for Word Representation}}.
\newblock \emph{Proceedings of the 2014 Conference on Empirical Methods in
  Natural Language Processing}, pages 1532--1543.

\bibitem[{Petrov et~al.(2010)Petrov, Chang, Ringgaard, and
  Alshawi}]{Petrov2010}
Slav Petrov, Pi-Chuan Chang, Michael Ringgaard, and Hiyan Alshawi. 2010.
\newblock Uptraining for accurate deterministic question parsing.
\newblock In \emph{Proceedings of the 2010 Conference on Empirical Methods in
  Natural Language Processing}, pages 705--713. Association for Computational
  Linguistics.

\bibitem[{Petrov and McDonald(2012)}]{Petrov2012}
Slav Petrov and Ryan McDonald. 2012.
\newblock \href {http://research.google.com/pubs/archive/38278.pdf} {{Overview
  of the 2012 shared task on parsing the web}}.
\newblock \emph{Notes of the First Workshop on Syntactic Analysis of
  Non-Canonical Language (SANCL)}, 59.

\bibitem[{Plank and van Noord(2011)}]{Plank2011}
Barbara Plank and Gertjan van Noord. 2011.
\newblock {Effective Measures of Domain Similarity for Parsing}.
\newblock \emph{Proceedings of the 49th Annual Meeting of the Association for
  Computational Linguistics: Human Language Technologies}, 1:1566--1576.

\bibitem[{Plank et~al.(2016)Plank, S{\o}gaard, and Goldberg}]{Plank2016a}
Barbara Plank, Anders S{\o}gaard, and Yoav Goldberg. 2016.
\newblock \href {http://arxiv.org/abs/1604.05529} {{Multilingual Part-of-Speech
  Tagging with Bidirectional Long Short-Term Memory Models and Auxiliary
  Loss}}.
\newblock In \emph{Proceedings of the 54th Annual Meeting of the Association
  for Computational Linguistics}.

\bibitem[{Rao(1982)}]{Rao1982}
C.~Radhakrishna Rao. 1982.
\newblock {Diversity and dissimilarity coefficients: a unified approach}.
\newblock \emph{Theoretical population biology}, 21(1):24--43.

\bibitem[{Rasmussen(2006)}]{Rasmussen2006}
Carl~Edward Rasmussen. 2006.
\newblock \emph{{Gaussian processes for machine learning}}.
\newblock MIT Press.

\bibitem[{Ravi et~al.(2008)Ravi, Knight, and Soricut}]{Ravi2008}
Sujith Ravi, Kevin Knight, and Radu Soricut. 2008.
\newblock Automatic prediction of parser accuracy.
\newblock In \emph{Proceedings of the Conference on Empirical Methods in
  Natural Language Processing}, pages 887--896. Association for Computational
  Linguistics.

\bibitem[{Remus(2012)}]{Remus2012}
Robert Remus. 2012.
\newblock {Domain adaptation using Domain Similarity- and Domain
  Complexity-based Instance Selection for Cross-Domain Sentiment Analysis}.
\newblock In \emph{IEEE ICDM SENTIRE-2012}.

\bibitem[{R{\'{e}}nyi(1961)}]{Renyi1961}
Alfr{\'{e}}d R{\'{e}}nyi. 1961.
\newblock \href {https://doi.org/10.1021/jp106846b} {{On measures of entropy
  and information}}.
\newblock In \emph{Proceedings of the fourth Berkeley symposium on mathematical
  statistics and probability}, volume~1.

\bibitem[{Rosenstein et~al.(2005)Rosenstein, Marx, Kaelbling, and
  Dietterich}]{Rosenstein2005}
Michael~T. Rosenstein, Zvika Marx, Leslie~Pack Kaelbling, and Thomas~G.
  Dietterich. 2005.
\newblock {To Transfer or Not To Transfer}.
\newblock \emph{NIPS Workshop on Inductive Transfer}.

\bibitem[{Ruder et~al.(2017)Ruder, Ghaffari, and Breslin}]{Ruder2017b}
Sebastian Ruder, Parsa Ghaffari, and John~G. Breslin. 2017.
\newblock \href {http://arxiv.org/abs/arXiv:1702.02426v1} {{Data Selection
  Strategies for Multi-Domain Sentiment Analysis}}.
\newblock In \emph{arXiv preprint arXiv:1702.02426}.

\bibitem[{Schnabel and Sch{\"{u}}tze(2014)}]{Schnabel2014}
Tobias Schnabel and Hinrich Sch{\"{u}}tze. 2014.
\newblock \href {http://acl2014.org/acl2014/Q14/pdf/Q14-1005.pdf} {{FLORS: Fast
  and Simple Domain Adaptation for Part-of-Speech Tagging}}.
\newblock \emph{TACL}, 2:15--26.

\bibitem[{Settles(2012)}]{Settles2012}
Burr Settles. 2012.
\newblock \emph{Active learning literature survey}.
\newblock Morgan and Claypool.

\bibitem[{Shannon(1948)}]{Shannon1948}
Claude~E Shannon. 1948.
\newblock \href {https://doi.org/10.1145/584091.584093} {{A mathematical theory
  of communication}}.
\newblock \emph{The Bell System Technical Journal}, 27(July 1928):379--423.

\bibitem[{Simpson(1949)}]{Simpson1949}
Edward~H. Simpson. 1949.
\newblock {Measurement of diversity}.
\newblock \emph{Nature}.

\bibitem[{Snoek et~al.(2012)Snoek, Larochelle, and Adams}]{Snoek2012}
Jasper Snoek, Hugo Larochelle, and Ryan~P. Adams. 2012.
\newblock \href {https://doi.org/2012arXiv1206.2944S} {{Practical Bayesian
  Optimization of Machine Learning Algorithms.}}
\newblock \emph{Neural Information Processing Systems Conference (NIPS 2012)}.

\bibitem[{S{\o}gaard(2011)}]{Sogaard2011}
Andres S{\o}gaard. 2011.
\newblock \href {http://dl.acm.org/citation.cfm?id=2002736.2002868} {{Data
  Point Selection for Cross-Language Adaptation of Dependency Parsers}}.
\newblock \emph{Proceedings of the 49th Annual Meeting of the Association for
  Computational Linguistics: Human Language Technologies (HLT '11): Short
  Papers}, pages 682--686.

\bibitem[{Tsvetkov et~al.(2016)Tsvetkov, Faruqui, Ling, and
  Dyer}]{Tsvetkov2016}
Yulia Tsvetkov, Manaal Faruqui, Wang Ling, and Chris Dyer. 2016.
\newblock \href {http://arxiv.org/abs/1605.03852} {{Learning the Curriculum
  with Bayesian Optimization for Task-Specific Word Representation Learning}}.
\newblock In \emph{Proceedings of the 54th Annual Meeting of the Association
  for Computational Linguistics (ACL 2016)}.

\bibitem[{{Van Asch} and Daelemans(2010)}]{VanAsch2010}
Vincent {Van Asch} and Walter Daelemans. 2010.
\newblock \href {http://eprints.pascal-network.org/archive/00007014/} {{Using
  Domain Similarity for Performance Estimation}}.
\newblock \emph{Computational Linguistics}, (July):31--36.

\bibitem[{van~der Wees et~al.(2017)van~der Wees, Bisazza, and
  Monz}]{vanderwees2017}
Marlies van~der Wees, Arianna Bisazza, and Christof Monz. 2017.
\newblock Dynamic data selection for neural machine translation.
\newblock In \emph{Proceedings of the 2017 Conference on Empirical Methods in
  Natural Language Processing}.

\bibitem[{Wu and Huang(2016)}]{Wu2016a}
Fangzhao Wu and Yongfeng Huang. 2016.
\newblock {Sentiment Domain Adaptation with Multiple Sources}.
\newblock \emph{Proceedings of the 54th Annual Meeting of the Association for
  Computational Linguistics (ACL 2016)}, pages 301--310.

\end{thebibliography}
\bibliographystyle{emnlp_natbib}

\end{document}